\documentclass{article}
\usepackage{ijcai20}

\usepackage{times}

\usepackage{soul}
\usepackage{url}
\usepackage[hidelinks]{hyperref}
\usepackage[utf8]{inputenc}
\usepackage[small]{caption}
\usepackage{graphicx}
\usepackage{amsmath}
\usepackage{booktabs}
\usepackage{amsmath,amsfonts,bm}

\def\eqref#1{equation~\ref{#1}}

\def\1{\bm{1}}

\def\mA{{\bm{A}}}

\def\mE{{\bm{E}}}

\def\mI{{\bm{I}}}

\def\mK{{\bm{K}}}
\def\mL{{\bm{L}}}

\def\mO{{\bm{O}}}

\def\mQ{{\bm{Q}}}

\def\mS{{\bm{S}}}

\def\mV{{\bm{V}}}
\def\mW{{\bm{W}}}

\DeclareMathAlphabet{\mathsfit}{\encodingdefault}{\sfdefault}{m}{sl}
\SetMathAlphabet{\mathsfit}{bold}{\encodingdefault}{\sfdefault}{bx}{n}

\def\emA{{A}}

\newcommand{\R}{\mathbb{R}}

\title{ET-USB: Transformer-Based Sequential Behavior Modeling for \\ Inbound Customer Service}

\author{
Ta-Chun Su$^1$\footnote{contact author at bgg@cathayholdings.com.tw}\and
Guan-Ying Chen$^1$
\affiliations
$^1$Cathay Financial Holdings Lab
\emails
\{bgg, amber\}@cathayholdings.com.tw
}

\begin{document}

\maketitle

\begin{abstract}
Deep learning models with attention mechanisms have achieved exceptional results for many tasks, including language tasks and recommendation systems. Whereas previous studies have emphasized allocation of phone agents, we focused on inbound call prediction for customer service. A common method of analyzing user history behaviors is to extract all types of aggregated feature over time, but that method may fail to detect users' behavioral sequences. Therefore, we created a new approach, ET-USB, that incorporates users' sequential and nonsequential features; we apply the powerful Transformer encoder, a self-attention network model, to capture the information underlying user behavior sequences. ET-USB is helpful in various business scenarios at Cathay Financial Holdings. We conducted experiments to test the proposed network structure's ability to process various dimensions of behavior data; the results suggest that ET-USB delivers results superior to those of delivered by other deep-learning models.
\end{abstract}

\section{Introduction}

Inbound call center customer service must deal with large numbers of customer complaints and queries; call center service quality is closely related to customer satisfaction, corporate brand image, and even profitability. Thus, many companies invest substantial resources to provide high-quality customer service via telephone calls, Internet communications, or chatbots. Several customer service applications are universal, including forecasting the number of phone calls in a given period, telemarketing various products to the customers, and predicting customers' call questions. \par 
In this paper, we focus on prediction of customers' questions for Cathay United Bank (a subsidiary of Taiwan's largest financial holding company, Cathay Financial Holdings), which receives nearly 1 million calls from customers every month. Usually, customers make phone calls only when they encounter operational problems. However, due to various types of miscommunication between call center agents and customers, the average call duration can exceed 3 minutes; this leads to high communication costs. Consequently, if we could predict the questions that customers are likely to ask, we could minimize call duration. Confirmation of customers' questions would be brief, and customer satisfaction would increase.

\begin{table*}[htb!]
\caption{Details of customer journey data.}
\begin{center}
\renewcommand{\arraystretch}{1.2}
\begin{tabular}{|c|c|c|c|c|c|c|c|}
\hline\hline
\textbf{customer\_id} & \textbf{action\_time} & \textbf{action\_type} & \textbf{channel\_type} & \textbf{object\_type} & \textbf{event\_type} & \textbf{attributes} \\ 
\hline
A123456 &1569859200 &40 &merchant\_nbr &credit\_card\_acct &cc\_transaction &\{key1: value1, ...\} \\
A123456 &1569911568 &41 &merchant\_nbr  &credit\_card\_acct &cc\_transaction &\{key1: value1, ...\} \\
A124789 &1569981819 &40 &merchant\_nbr  &credit\_card\_acct &cc\_transaction &\{key1: value1, ...\}\\
\hline\hline
\end{tabular}
\label{table1}
\end{center} 
\end{table*}

Inspired by the success of Vaswani et al. with their attention mechanism for machine translation tasks in natural language processing (NLP) \cite{Bahdanau2015}, we propose a novel neural network, ET-USB, which utilizes the encoder in the transformer of Vaswani et al. \cite{Vaswani2017}, to learn highly effective representations of users' sequential behaviors. Our model utilizes the self-attention mechanism to capture the most important dependencies among behavior elements in a sequence; ET-USB can output excellent predictions regarding inbound customer calls. We found that customer calls were typically triggered by difficulties encountered with various channels or product services. In other words, customer interactions with each channel, product, or service are crucial to analyzing inbound call problems. Our company uses the HIPPO\ framework to collect user data from various subsidiary databases; HIPPO\ is an event-driven information integration framework for aiding workflow. In our system, a customer journey event (CJE)\ is a particular set of user behavior data acquired from over 30 channels. CJE times can be analyzed to determine users' behavioral sequences, which can be used for various scenarios, including solving inbound call problems. \par

We conducted experiments on the granularity of the behavior CJEs. In one coupon redemption example, each item in a sequence was embedded without complex feature engineering for easy interpretation; items included "click," "redeem,"\ and "gift-sending." (Similar to Item2Vec \cite{Barkan2017} approach). However, in other scenarios, describing a behavior with a simple approach is difficult. Assume that two sentences about viewing a movie refer to similar events but use different terms. Consider the sentences, "I watched \textit{Avengers: Endgame} in the movie theater" and "I watched \textit{Avengers: Endgame} on AMC during the evening." These sentences illustrate how difficult it can be to create a universal approach for effective description of behaviors. Therefore, we introduce a statistical method to improve the granularity with which behavior can be described. Therefore, we used offline experiments to demonstrate the effectiveness of our proposed approaches. When we deployed the model online at Cathay United Bank, it accurately predicted inbound calls for customer service.

\section{Related Work}

In recent years, unsupervised pretrained models have dominated the field with embedding approaches; in typical systems, preliminary results are obtained rapidly then fed into fine-tuning models later. During pretraining, an encoder neural network model is trained using large-scale unlabeled data to learn word embeddings; parameters are then fine-tuned with labeled data related to downstream tasks. For example, for text classification of Wikipedia articles, each word in a sentence is embedded into a low-dimensional spaces within vector representations that are learned from large quantities of unstructured textual data. Then, results are fed into fully connected layers to predict the categories to which each article belongs. Among such methods, Word2vec is the most common \cite{Mikolov2013}. \par

Because of the effectiveness of Word2vec, several non-NLP applications have used pretrained vectors as inputs to solve related tasks. For example, Item2Vec \cite{Barkan2017} uses the concept of Word2Vec and produces embeddings for items in a latent space. The method can infer item-to-item relations even when user information is unavailable. Embedding vectors contain more information than one-hot encoding; viable methods include everything2Vec, node2vec \cite{Grover2016}, and graph2vec \cite{Narayanan2017}. These embedding vectors can be further fed into a deep learning (DL) architecture.

In current architectures, attention mechanisms learn to focus on only the most essential parts of targets. Attention mechanisms are effective in tasks such as computer vision and machine translation. Vanilla attention mechanisms have been integrated into DL algorithms (e.g., RNN and CNN) \cite{Sinha2018} \cite{Zhou2016}, but they present problems of long-range dependency and inability to be parallelized; thus, these approaches are inefficient in some scenarios. Self-attention \cite{Vaswani2017} was introduced by Vaswani et al. BERT \cite{Devlin2019} uses the concept of self-attention with the Transformer model and demonstrated state-of-the-art performance and efficiency on the General Language Understanding Evaluation Leaderboard, which had previously been dominated by recurrent and convolutional neural network approaches. The Transformer model relies heavily on the self-attention mechanism to capture complex structures in sentences. Many researchers have applied the aforementioned approaches to DL-based recommendation systems. Transformer was applied in DIN\cite{GuoruiZhou2018}, an adaptive method for learning user interests from historical behaviors with respect to certain advertising techniques. Atrank\cite{ChangZhou2018} used an attention-based user behavior modeling framework for recommendation tasks. Alibaba also recently introduced a Transformer-based recommendation system \cite{Chen2019}.

\begin{figure*}[t]
        \begin{center}
        \includegraphics[scale=0.26]{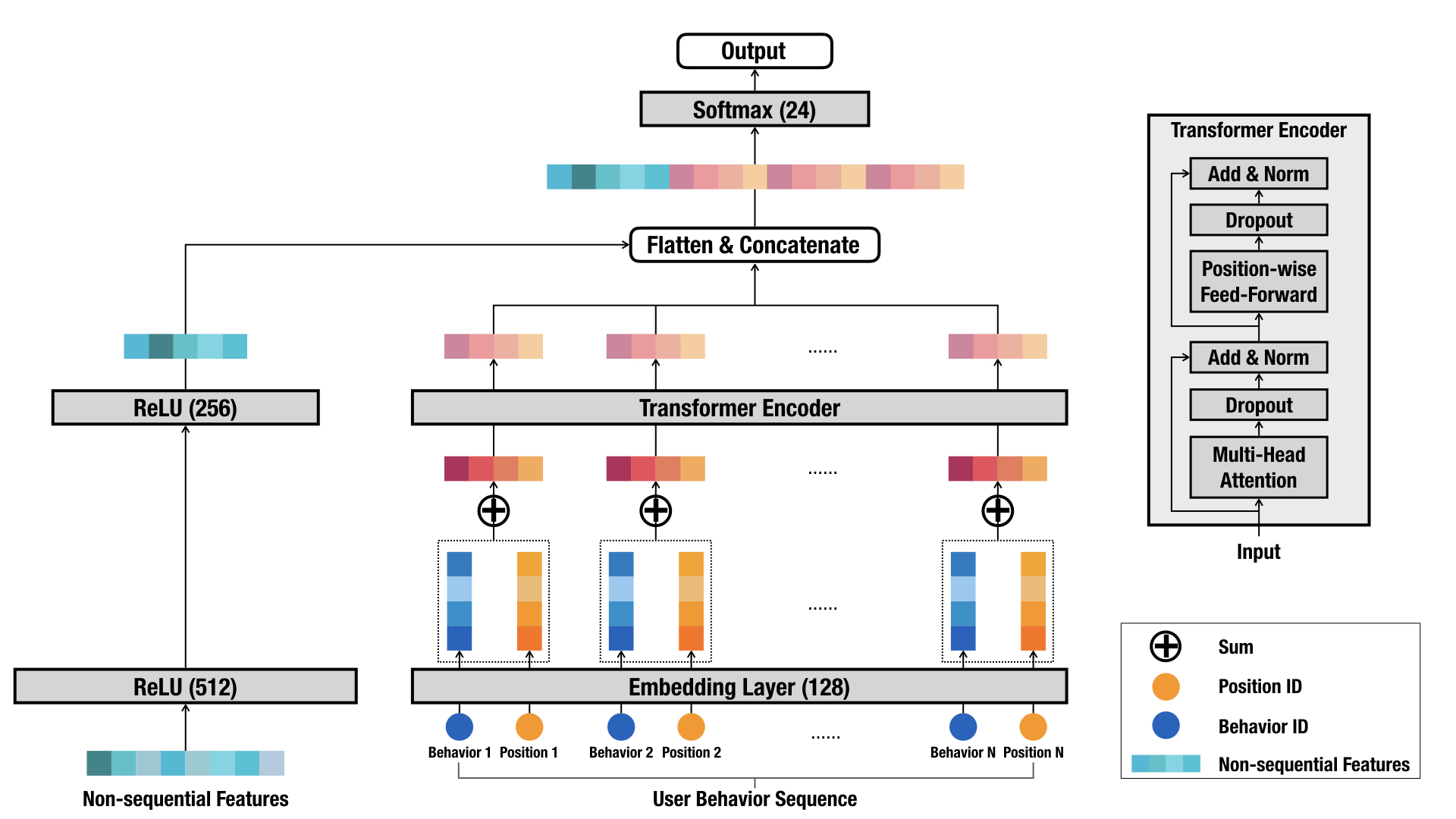}
        \caption{\label{Figure1}Architecture of ET-USB. The model separately considers sequential features and nonsequential features. Specifically, the model embeds sequential features in low-dimensional vectors. The self-attention mechanism operates inside encoder layers to capture the relations among behaviors in a sequence to more deeply learn how to represent them. The system concatenates the output of nonsequential features, which are processed through two fully connected layers; that result is fed into the final layer, which executes the softmax function to generate the final output.}
        \end{center}
\end{figure*} 

\section{Modeling Framework}
\subsection{Description of Customer Journey Data}

To delineate outline customer behavior, we integrate the CJEs occurring through various channels into customer journey data. Table \ref{table1} provides an example of the raw data of a CJE\ for a credit card transaction. Various dimensions of user transaction behaviors are recorded, including customer ID, action time, and attributes (including channel type, card type, and other external domain information). For any specific CJE, the system can access information regarding when a behavior happened, over which channel (e.g., online/offline store), and with which credit card. The company's privacy policy stipulates that we do not disclose any customer journey data in this paper. In sum, over 20 elements of attributes are collected from each channel for analysis.

\subsection{Model Architecture}
Our model architecture is shown in Figure \ref{Figure1}. We divide our behavior features into two categories: nonsequential and sequential features. In the subsequent section, we describe information extraction from sequential behavior features; nonsequential features are used only for concatenation with information in the final layer.

\subsubsection{Embedding Layer}
We embed all behaviors in behavior sequences into fixed-size low-dimensional vectors. Given a vocabulary $V$ of behaviors, we create an embedding matrix $\mE \in \R^{V \times d}$, where $d$ is the dimension size. $\mE$ is referred to as the behavior lookup table. The size of embeddings $d$ is often set between 100 and 1000 \cite{Bahdanau2015}. Here we set $d = 128$ and adopt the Xavier initialization method. \par 
The position of each behavior must be represented; otherwise identical behaviors at different positions would have the same output. The positions of sequential behaviors might have different meanings and latent information in various behavior sequences. We add position as an input feature of each item in the input layer before projection as a low-dimensional vector.

\subsubsection{Transformer Encoder}
\paragraph{Self-Attention Layer}
Consider the encoder architecture of a multilayer Transformer \cite{Vaswani2017}. The encoder consists of stacked encoder layers, each containing two sublayers, namely a multihead self-attention layer and a feed-forward network. The multihead mechanism runs through a scaled dot-product attention function, which can be formulated by querying a dictionary entry with key value pairs \cite{Miller2016}. The self-attention input consists of a query $\mQ \in {\R^{l\times d}}$, a key $\mK\in {\R^{l\times d}}$, and a value $\mV \in {\R^{l\times d}}$, where $l$ is the length of the input sentence, and $d$ is the dimension of embedding for the query, key, and value. For subsequent layers, $\mQ$, $\mK$, $\mV$ are derived from the output of the previous layer. The scaled dot-product attention \cite{Vaswani2017} is defined as follows: self-attention allows capturing the dependencies between representation pairs without regard to any distance between sequences. 

\begin{equation}
    Attention(\mQ,\mK,\mV) = softmax(\frac{\mQ \mK^{T}}{\sqrt{d}})\cdot \mV = \mA \cdot \mV
\end{equation}

The output represents the product of the attention weights $\mA$ and vector $\mV$, where $\mA \in \R^{l\times l}$. The attention weights $\emA_{i,j}$ quantify the relevance of the $i$th key-value pair with respect to the $j$th query to the generated output \cite{Bahdanau2015}.

Linearly projecting queries, keys, and values $h$ times with different representation subspaces at different positions is beneficial. Specifically, multihead attention first linearly projects input layer $\mI$ onto $h$ subspaces with different linear projections, and then executes the attention function in parallel, yielding output representations, which are concatenated and reprojected.

\begin{align}
    \mO^{head} & = MultiHead(\mI) \nonumber\\
    & = Concat(\mO^{head}_{1}, \mO^{head}_{2}, ..., \mO^{head}_{h})\mW^{O}
\end{align}

\begin{align}
    \mO^{head}_{i}=Attention(\mI\mW_{i}^{Q}, \mI\mW_{i}^{K}, \mI\mW_{i}^{V})
\end{align}

where the projection matrices for each head $\mW^{Q}_{i}, \mW^{K}_{i}, \mW^{V}_{i} \in \R^{d\times \frac{d}{h}}$, and $\mW^{O}\in \R^{d\times d}$ are learnable parameters, and \textit{h} is the number of heads.

\begin{table*}[htb!]
\caption{Comparison of training and prediction data sets. We adopted a validation set, which constituted 10\% of the training data set, for model evaluation.}
\begin{center}
\renewcommand{\arraystretch}{2}
\begin{tabular}{ |c|c|c|c| } 
\hline
Customer Journey Data  & \# Users  & 
  \# Event Instances & \# Questions Categories\\
  \hline
     Training Data sets  &971,904 &16,510,346 &24 \\
   \hline
    Prediction Data sets &4,635,236 &99,678,852 &24  \\
  \hline
\end{tabular}
\label{table2}
\end{center}
\end{table*} 
      
\paragraph{Feed-Forward Network Layer}
Because the self-attention sublayer is mainly based on linear projections. To enhance the model with nonlinearity, we apply a feed-forward network after the self-attention layer. This consists of two linear transformations with scaled exponential linear unit (SeLU) activation in between. The outputs of the feed-forward layers are formulated as follows:

\begin{align}
     \mO^{out} = SELU(\mO^{head}\mW^{(1)}+b^{(1)})\mW^{(2)}+b^{(2)}
\end{align}

Next, we employ a residual connection \cite{He2015} by applying dropout to the output of each sublayer before it is added to the sublayer input, followed by layer normalization \cite{Ba2016}.

\begin{align}
     \mL = LayerNorm(\mO^{head} + Dropout(\mO^{out}))
\end{align}

\paragraph{Stacking Encoder Layer}
After the first self-attention encoder layer, the system aggregates all the previous behavior embeddings. Because stacking the previous encoder layers $L $ may be helpful for learning complex behavior transition patterns, we stack the encoder layers, including the self-attention layer and feed-forward network. The $b$th block is defined as follows:
\begin{align}
    \mS^{(b)} = SA(\mL^{(b-1)}) \\
    \mL^{(b)} = FFN(\mS^{(b)})  
\end{align}
where $SA$ is the self-attention layer, $FFN$ is the feed-forward network, and $b \geq$ 1. Specifically, the performance of $b$ depends on the scenario. The experiments described in this paper demonstrate that $b =$1 is optimal.

\paragraph{Output Layer}
We use two fully connected layers to learn the interactions among nonsequential features; next, we concatenate their output with the output of the Transformer layer. Together, this forms dense representation vectors. 

\subsection{Objective Function}
To predict which the category of question a customer asks, we model the topic as a multiclass classification problem; thus, we use the softmax function as the output unit. The objective function used in the model provides the category cross-entropy as follows:
\begin{align}
    Loss =  -\frac{1}{N} \sum_{i=1}^{N}\sum_{j=1}^{L}y_{ij}log(\hat{p}_{ij}) \\
   \hat{p}_{ij} = \frac{\exp^{f_{j}(x_{i})}}{\sum_{j\prime=1}^{L}\exp^{f_{j\prime}(x_{i})}}
\end{align}
where $N$ represents the size of training set, $y_{ij}$ corresponds to the $j\prime$th element of the one-hot encoded label of the sample $x_{i}$, and $f_{j}$ denotes the $j\prime$th element of $f$. 

\section{Experiments}

In this section, we present our experiments in detail, including data description, data processing, model comparison, and the corresponding analysis. We evaluate the proposed approach, ET-USB, according to its performance in a multiclass classification task. Our experiments demonstrate that ET-USB outperforms prestigious methods at NLP tasks.

\subsection{Setup}
In our experiments, we implemented the models using TensorFlow. All of the results in this paper were processed in 2 hours or less by a graphics processing unit with training data sets. We trained all of the models in the same computation environment, which featured an NVIDIA Tesla V100 graphics processing unit.

\subsubsection{Training and Prediction Data Sets}
To form the training and prediction data sets, we cooperated with business units to decide the business logic of relevant data sets; we excluded meaningless information to denoise the data. Furthermore, because certain call questions tend to occur in specific months, the training data set consisted of inbound calls from the last four months and months with specific call questions. Most importantly, we predicted the categories of only those call questions related to credit cards, which accounted for a total of 24 targets. The statistics of the data sets are shown in Table \ref{table2}.

\begin{table*}[htb!]
\caption{Adjustment of behavior granularity by cumulative percentage.}
\begin{center}
\renewcommand{\arraystretch}{2}
\begin{tabular}{|c|c|c|c|c|}
\hline
\textbf{event} & \textbf{txn\_currency\_code} & \textbf{count} & \textbf{event\_cnt} & \textbf{event\_percent} \\ 
\hline
creditcard\_transaction &TWD &72,761,143 &75,177,802 &0.967854 \\ 
\hline
creditcard\_transaction &others &2,416,659 & 75,177,802 &0.032146 \\ 
\hline
\end{tabular}
\label{table3}
\end{center}
\end{table*} 

\subsection{Data Preprocessing}
\subsubsection{Granularity of CJEs}
\begin{figure}[htb!]
    \begin{center}
    \includegraphics[scale=0.26]{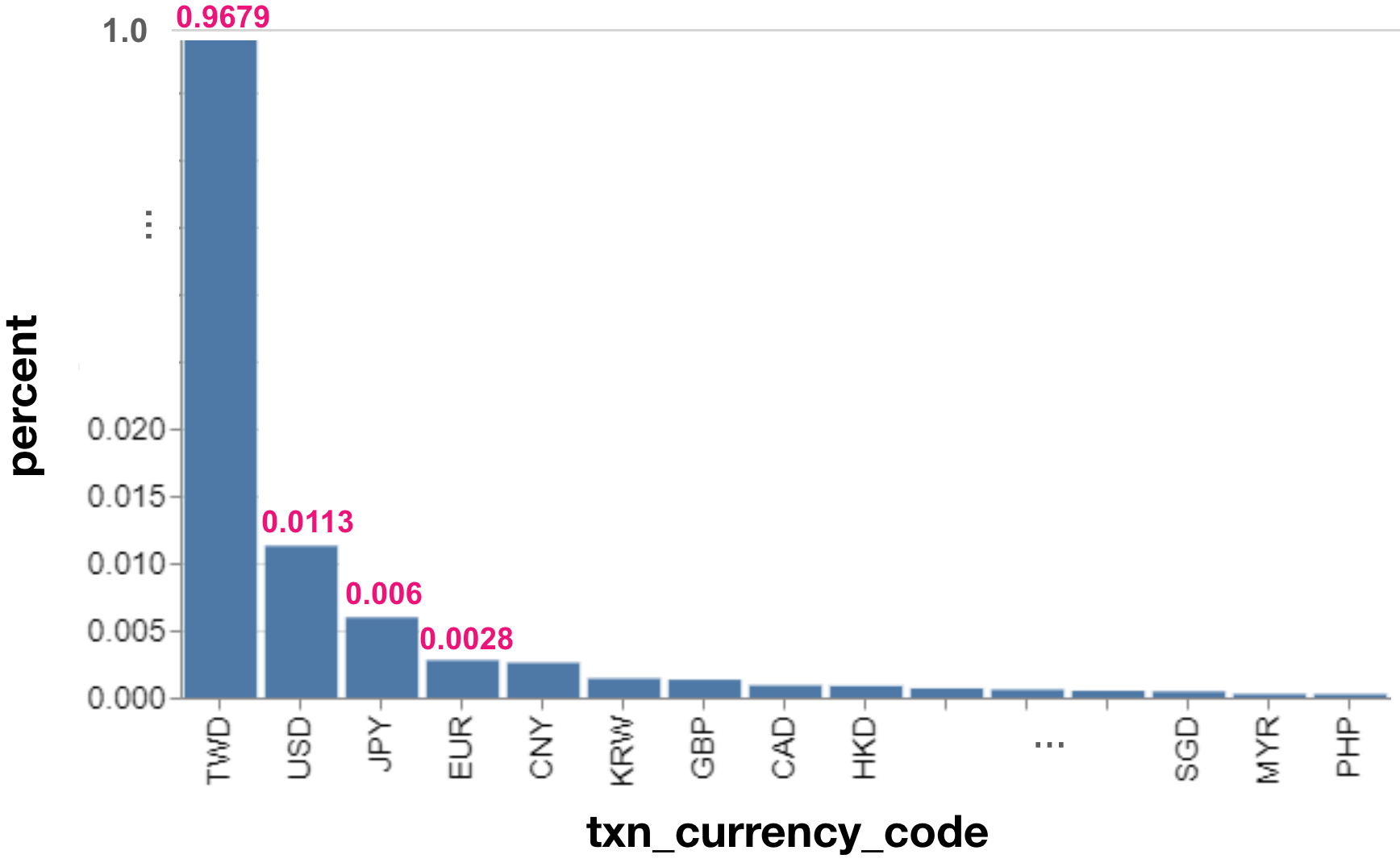}
    \caption{\label{Figure2}The Data Distribution of Currency Code}
\end{center}
\end{figure}

The raw data of CJEs contain rich information about behavior, which has both advantages and disadvantages. The company can parse these momentary events to understand each customer in detail. Sufficient granularity is necessary to fully understand the nature of the information. However, if the information is examined with excessive granularity, the noise can overwhelm the signal. However, use of extremely rough granularity, as presented in Table \ref{table1}, would render extraction of information from user behavior sequences challenging. In practice, the data scientist must discover the proper granularity to describe the customer behavior event and reveal the commonalities among all customers' behaviors. This concept is similar to stemming and lemmatization in NLP. 
We propose a statistical method to reduce behavior granularity. We utilize feature transformation methods for both numerical and categorical features and adjust the behavior granularity to suit the cumulative percentages and financial domain knowledge. For instance, after discretizing continuous features into categorical ones, we sort each value in descending order. Thus, we may calculate the cumulative percentages of the transformed categorical features. According to some specific threshold value (e.g., 90 percent), we retain the top \textit{N} distinct values and transform the remaining values into the "other" type. Figure \ref{Figure2} presents the original data distribution, a long-tailed distribution, of currency codes; over 100 currencies are used for credit card transactions. Nevertheless, the New Taiwan dollar (TWD) is used in an overwhelming majority of transactions. Therefore, we reduce the currency codes to classify currency as either TWD or other currencies to simplify our search for the commonalities among customer behaviors. The result is presented in Table \ref{table3}.

\subsubsection{Feature Extraction}
\begin{figure}[htb!]
    \begin{center}
    \includegraphics[scale=0.13]{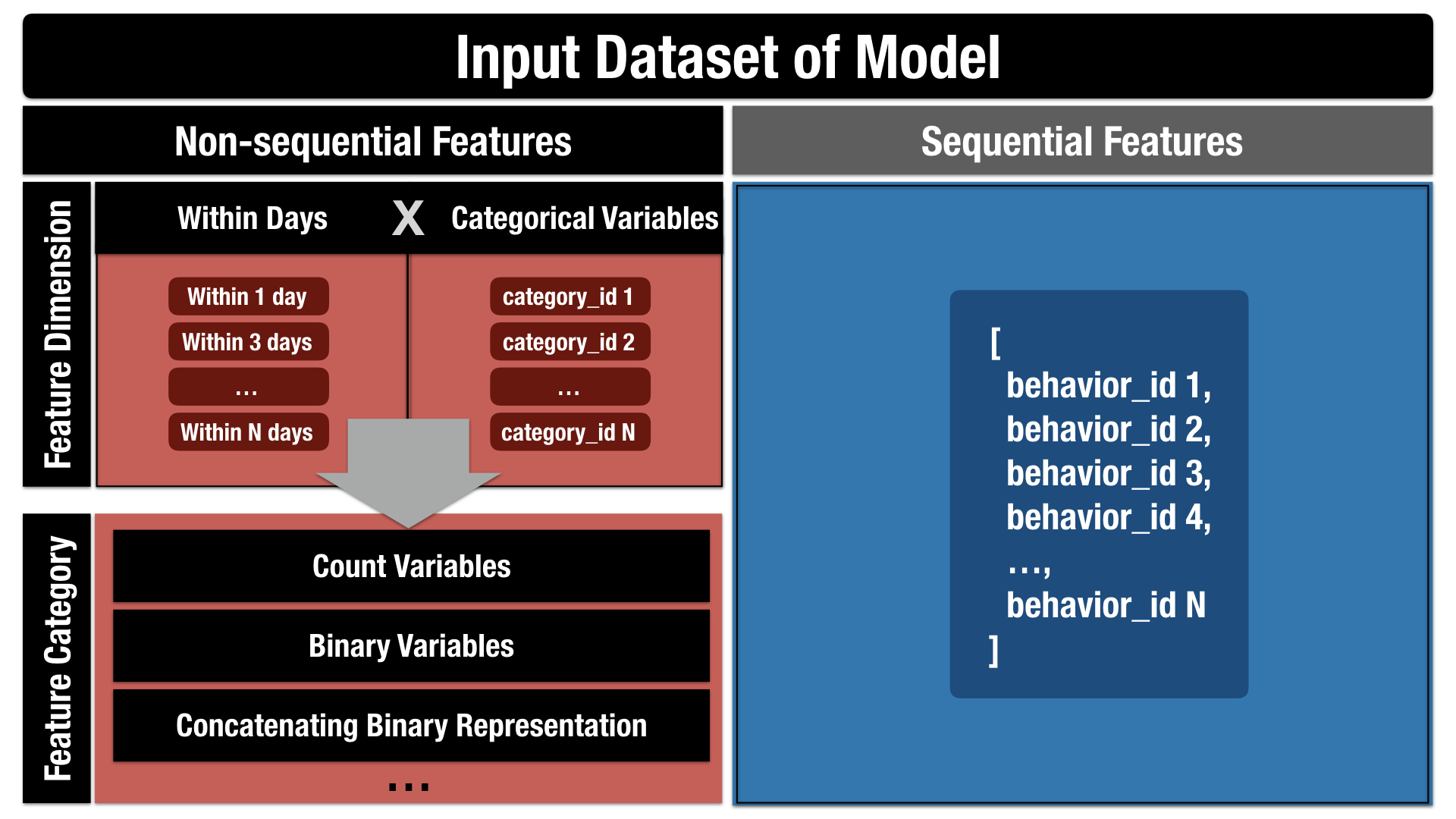}
    \caption{\label{Figure3}Input data set of the model.}
\end{center}
\end{figure}

We emphasize both sequential and nonsequential features extracted from CJEs. Here, we demonstrate some nonsequential features for simplicity. We do not use demographic features, such as gender or age. We believe that inbound calls are unrelated to these features. We care about the timing of the latest behavior event, specifically whether this event occurred within $e_{i}$ days, where $e_{i} \in \{1, ..., N\}$ and $N$ is a positive integer. Furthermore, we transform categorical variables using one-hot encoding and combine features in different dimensions. For example, we expand the unique value in each categorical field by cross-referencing with time features. We also count the unique values. The input data set of the model is shown in Figure \ref{Figure3}.

\begin{figure*}[t]
        \begin{center}
        \includegraphics[scale=0.26]{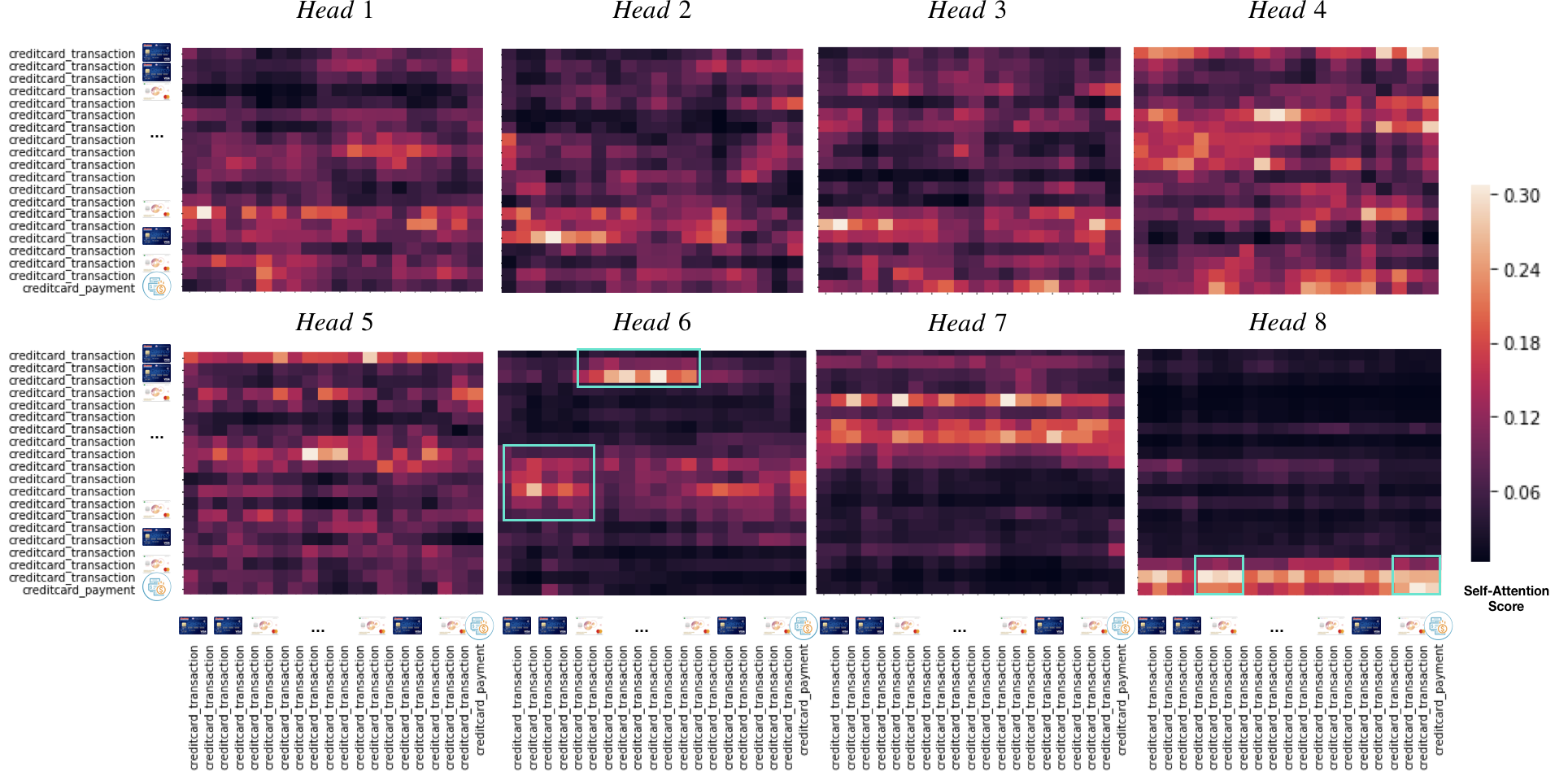}
        \caption{\label{Figure4}Heatmap of self-Attention scores using the Transformer encoder. We employ an $n$ = 1 for the encoder layer.}
        \end{center}
\end{figure*} 

\subsection{Experiment Results}
This subsection provides the experiment results of the baseline model and our proposed method. We employed a batch size of 128, learning rates of 3e-4, one stacking layer, and three epochs for our tasks. The reason that we used only one stacking layer is that we found that systems with more stacking layers did not perform as well as systems with one layer, as mentioned in \cite{Chen2019}. One behavior in a sequence is probably not as informative as one word in a sentence. Therefore, a small number of layers should be sufficient. The results are presented in Table \ref{table4}, which illustrates the advantages of ET-USB over other approaches. Compared with base models without sequential behavior features, we can see the advantages of those features in the subsequent models. 
\begin{table}[htb!]
\caption{Performance levels of various models on training data sets, with map@3 and accuracy as the metrics.}
\begin{center}
\renewcommand{\arraystretch}{1.2}
\begin{tabular}{c|cc}
  \hline\hline
  & \bf{Map@3} & \bf{Accuracy}   \\
  \hline
     DNN (w/o Seq) &0.2345 &0.1826  \\
     CNN   &0.2542 &0.2468 \\
     Bi-LSTM  &0.2527 &0.2472 \\
     Bi-LSTM+Att  &0.2546 &0.2491 \\
     ET-USB  &\bf{0.2657} &\bf{0.2562} \\
     \hline\hline
 \end{tabular}  
    \label{table4}
\end{center}    
\end{table} 
Comparing the effectiveness of incorporating behavior sequence information in different models reveals that ET-USB using self-attention is significantly more effective. 
\par
We can explain the effects of the self-attention mechanism in our model by visualizing a user's behavior sequence with customer journey data. Figure \ref{Figure4} illustrates how the self-attention scores distribute behaviors to different heads. The behavior in coordinates is ordered by the happening timeline. We use a heatmap matrix to represent the self-attention scores between any two behaviors in several different semantic spaces. Notably, this matrix is normalized through row-based softmax. Consider the user in Figure \ref{Figure4}, who called regarding a credit-card bill. We present the case at the roughest level of granularity to explain the behavior in terms of coordinates. Subsequently, we explain the phenomenon using a more detailed level of granularity. Specifically, this user bought a cell-phone with a COSTCO card, bought a plane ticket using a KOKO Combo card, and then executed a series of behaviors; eventually he tried to pay the credit card bill at a convenience store. He was confused when he attempted to pay the credit card bill, and thus he made a inbound call, which was classified as "credit-card bill." Here we can see how the model works. Different semantic spaces may focus on different behaviors, and the heatmaps vary greatly for some latent spaces. For example, $Heads \ 6, 7, 8$ are different from the other five heads. In the other five heads, the relative trends of the attention score vectors are similar, regardless of strength. These heads mostly consider the overall centrality of the behavior among them. Regarding $Head \  6$ and $Head \ 8$, we found that high scores tended to form in a dense situation. In $Head \  6$, early behaviors formed a group of credit card transaction behaviors, with very high attention scores. Those spaces probably consider relationships among similar qualities of transaction behavior of a certain granularity. However, $Head \ 8$ had a high attention score for the latest behaviors; that is, this space focused on credit card payment behaviors. From these heads, we conclude that the self-attention mechanism greatly influenced which user behaviors were distributed in which semantic spaces.

\section{Conclusion}
In this paper, we propose an innovative sequential behavior modeling method, called ET-USB, for customer inbound call prediction. ET-USB applies self-attention to capture behavior relations. Additionally, we provide a statistical approach to adjust the granularity of customer behavior representations. Experimental results reveal that ET-USB outperforms both a model that does not incorporate sequential behaviors and also popular sequential behavior models in the field of NLP. The results also reveal the self-attention score distributions of behaviors in various heads. The model architecture has already been deployed in a production environment at Cathay United Bank to help phone agents accurately predict customer questions, thereby reducing communication costs and improving the efficiency of human resource allocation for customer service. In the future, we will investigate each behavior embedding in different channels of customer journey data that may contain latent information regarding users' sequential behaviors.

\bibliographystyle{named}
\bibliography{ijcai20}

\end{document}